\documentclass{article}
\usepackage{spconf,amsmath,graphicx}
\usepackage{graphicx}
\usepackage{microtype}
\usepackage{amsmath,amssymb,amsfonts}
\usepackage{algorithm,algorithmic}
\usepackage{subfigure}
\usepackage{booktabs} 
\usepackage{comment}
\usepackage{theorem}
\usepackage{amsmath,amssymb} 
\usepackage{color}

\def\ab{{\mathbf a}}
\def\db{{\mathbf d}}
\def\xb{{\mathbf x}}
\def\cb{{\mathbf c}}
\def\zb{{\mathbf z}}

\def\Ab{{\mathcal A}}
\def\Ib{{\mathbf I}}
\def\Pb{{\mathcal P}}
\def\Mb{{\mathcal M}}
\def\Fb{{\mathbf F}}
\def\Db{{\mathbf D}}
\def\Qb{{\mathbf Q}}
\def\Wb{{\mathbf W}}

\def\Zb{{\mathbf Z}}
\def\Ub{{\mathbf U}}
\newcommand{\argmind}[2]{\ensuremath{\underset{\substack{{#1}}}%
{\mathrm{argmin}}\;\;#2 }}

\newtheorem{theorem}{Theorem}[section]

\title{Deep Transform and Metric Learning Networks}
\name{Wen Tang$^*$, Emilie Chouzenoux$^\dagger$, Jean-Christophe Pesquet$^\dagger$,~and Hamid Krim$^*$\thanks{$^*$This research work was generously supported in part by the U.S. Army Research Office under agreement W911NF1910202.}}
\address{$^*$Department
of Electrical and Computer Engineering, North Carolina State University, Raleigh, NC, 27606\\
$^\dagger$ Universit\'e Paris-Saclay,  Inria, CentraleSup\'elec,
Center for Visual Computing, 91190 Gif sur Yvette, France\\
\small{E-mail: \{wtang6,~ahk\}@ncsu.edu, $^\dagger$\{emilie.chouzenoux,~jean-christophe.pesquet\}@centralesupelec.fr}}

\begin{document}
%
\maketitle
\begin{abstract}
Based on its great successes in inference and denosing tasks, Dictionary Learning (DL) and its related sparse optimization formulations have garnered a lot of research interest. While most solutions have focused on single layer dictionaries, the recently improved Deep DL methods have also fallen short on a number of issues. We hence propose a novel Deep DL approach where each DL layer can be formulated and solved as a combination of one linear layer and a Recurrent Neural Network, where the RNN is flexibly regraded as a layer-associated learned metric. Our proposed work unveils new insights between the Neural Networks and Deep DL, and provides a novel, efficient and competitive approach to jointly learn the deep transforms and metrics. Extensive experiments are carried out to demonstrate that the proposed method can not only outperform existing Deep DL, but also state-of-the-art generic Convolutional Neural Networks.
\end{abstract}
\section{Introduction}
\label{introduction}
Dictionary Learning/Sparse Coding 
has been successfully applied for solving various inference tasks, such as image denoising \cite{denoising}, image restoration \cite{imagerestoration}, image super-resolution \cite{superresolution}, audio processing \cite{audioprocessing}, and image classification \cite{imageclassification}. 

Synthesis Dictionary Learning (SDL) is the early primary approach in this area. Then, Analysis Dictionary Learning (ADL)/Transform Learning, the dual problem of SDL, received more attention.
DL based methods \cite{ksvd,sksvdadl,tang2018structured,tang2019convolution,tang2019analysis} for image classification usually focus on learning one-layer dictionary and its associated sparse representation. 
Since DL methods need to simultaneously learn the dictionary and its associated representation, thanks to a sparsity promoting regularization, they raise challenging numerical issues. To address such challenges, a number of dictionary learning solvers have been developed, including K-SVD \cite{ksvd} and Fast Iterative Shrinkage-thresholding Algorithm (FISTA) \cite{fista} schemes, for instance. 
Although such alternating minimization methods provide practical solutions for DL, they remain prone to limitations and have a relatively high computational cost. 
To overcome such computational difficulties, differentiable programming has also been developed, to take advantage of the efficiency of neural networks, such as LISTA \cite{LISTA} and Sparse LSTM \cite{zhou2018sc2net}. 

Although the aforementioned differentiable programming methods are efficient at solving a single-layer DL problem, the latter formulation still does not lead to the best performance in image classification tasks. 
With the fast development of deep learning, Deep Dictionary Learning (DDL) methods \cite{tariyal2016deep,mahdizadehaghdam2017image} have thus come into play. \cite{mahdizadehaghdam2019deep} deeply stacks SDLs to classify images and achieves promising performance. Also, a deep model for ADL followed by a SDL is developed for image super-resolution in \cite{huang0418_DDSR}. Unsupervised DDL approaches have also been proposed, with interesting results \cite{Gupta,Maggu2018}.

However, to the best of our knowledge, there is no method for training such kind of end-to-end deep dictionary models in a both fast and reliable manner.
This work aims at ensuring the discriminative ability of single-layer DL while benefiting from the efficiency of end-to-end models. To this end, we propose a novel differentiable programming method, namely, Deep Transform and Metric Learning. To further benefit from the efficiency of DDL models, the resulting structures are stacked into a deep network,
leading to a so-called DeTraMe Network (DeTraMe-Net). Our new approach not only increases the discrimination capabilities of DL, but also brings the flexibility to construct different DDL or Deep Neural Network (DNN) architectures. It also allows to overcome the roadblocks raised by initialization and gradient propagation issues arising in standard DDL methods.
Although the authors of \cite{wang2015deep} and \cite{liu2018dictionary} also used a CNN followed by an RNN for respectively solving super-resolution and sense recognition tasks, they directly used LISTA in their model. In turn, our method actually solves the same problem as LISTA. 


Our main contributions are summarized below: 1) We theoretically transform one-layer dictionary learning into a transform learning and Q-Metric learning, and derive the conversion of DDL into DeTraMe-Net. 2) Such joint transform learning and Q-Metric learning are 
easily implemented into a combined linear layer and a RNN. A convolutional layer can be chosen for the linear part, and the RNN can also be simplified into a Convolutional-RNN. To the best of our knowledge, this is the first work which establishes an insightful bridge between DDL methods and the combination of Forward Neural Networks (FNNs) and RNNs. 3) The transform and Q-Metric learning use two independent variables, one for the metric and the other for the pseudo-inverse of the dictionary. This allows us to make a link with conventional SDL while introducing more discriminative power and allowing faster learning procedures than the original DL. 4) Q-Metric can also be viewed as a parametric non-separable nonlinear activation function, while in current neural network architectures, very few non-separable nonlinear operators are used. 
As a module in a neural network, it can be flexibly inserted into any network architectures to easily construct DL layer. 5) The proposed DeTraMe-Net is demonstrated not only to improve the discrimination power of DDL, but also to achieve a better performance than the state-of-the-art CNNs.

The paper is organized as follows: In Section \ref{sec:related}, we introduce the required background material. We derive the theoretical basis for our novel approach and its algorithmic solution in Section \ref{sec:DeTraMe-Net}. Substantiating experimental results and evaluations are presented in Section \ref{sec:experiments}. Finally, we provide some concluding remarks in Section \ref{sec:conclusion}.

\section{Preliminaries}
\label{sec:related}

\subsection{Notation}
Uppercase and lowercase bold letters respectively denote matrices and vectors throughout the paper. The transpose and inverse of matrices are respectively represented by the superscripts $\null^\top$ and $\null^{-1}$.
The identity matrix is denoted by $\mathbf{I}$. The lowercase $a_{i,j}$ represents the element in the $i^{th}$ row and $j^{th}$ column of matrix $\mathbf{A}$, and $a_i$ represents the $i^{th}$ component of vector $\mathbf{a}$.
\subsection{Deep Dictionary Learning}
An efficient DDL approach \cite{mahdizadehaghdam2019deep} consists of computing $\mathbf{\hat{y}=\varphi (C x^{(s)})}$
where
\begin{equation} \label{equ:ddl-1}
    \xb^{(s)}= \Pb^{(s)}\circ \Mb_{\Db^{(s)}} \circ \dots \circ \Pb^{(1)}\circ \Mb_{\Db^{(1)}}(\xb^{(0)}),
\end{equation}
and $\mathbf{\hat{y}}$ denotes the estimated label, $\mathbf{C}$ is the classifier matrix, $\mathbf{\varphi}$ is a nonlinear function, and $\circ$ denotes the composition of operators. For every layer $r \in \{1,\dots, s\},~\Pb^{(r)}$ is a reshaping operator, which is a tall matrix. Moreover, $\Mb_{\Db^{(r)}}$ is a nonlinear operator computing a sparse representation within a synthesis dictionary matrix $\Db^{(r)}$. More precisely, for a given matrix $\Db^{(r)} \in \mathbb{R}^{m_r \times k_r}$,
\begin{equation} \label{equ:ddl-main}
\begin{split}
        &
         \Mb_{\Db^{(r)}}:~\mathbb{R}^{m_r} \to \mathbb{R}^{k_r}, ~\xb \mapsto \argmind{\mathbf{a}\in \mathbb{R}^{k_r}}~\mathcal{L}^R(\Db^{(r)},\mathbf{a},\xb),
         \\
        &
        \mathcal{L}^R(\Db^{(r)},\mathbf{a},\xb)
        =\frac{1}{2}\|\xb-\Db^{(r)}\ab\|^2_F+\lambda \psi_r(\ab)+\frac{\alpha}{2}\|\ab\|_2^2\\
        &\qquad\qquad\qquad\quad+(\db^{(r)})^\top \ab, 
\end{split}
\end{equation}
where $(\lambda,\alpha) \in (0,+\infty)^2$, $\db^{(r)}\in \mathbb{R}^{k_r}$, and $\psi_r$ is a function in $\Gamma_0(\mathbb{R}^{k_r})$, the class of proper lower semicontinuous convex functions from $\mathbb{R}^{k_r}$ to $(-\infty,+\infty]$. A simple choice consists in setting $\db^{(r)}$ to zero, while 
adopting the following specific form for $\psi_r:$ 
    $\psi_r =\|\cdot\|_1+\iota_{[0,+\infty)^{k_r}},$
where $\iota_S$ denotes the indicator function of a set $S$ (equal to zero in $S$ and $+\infty$ otherwise). Note that Eq.~(\ref{equ:ddl-main}) corresponds to the minimization of a strongly convex function (w.r.t. $\ab$), which thus admits a unique minimizer, so making the operator $\Mb_{\Db^{(r)}}$ properly defined.

\section{Deep Transform and Metric Learning}
\label{sec:DeTraMe-Net}

\subsection{Proximal interpretation}

In the following, the superscript denoting the layer $r$ has been omitted for simplicity.
We have then the following result \cite{tang2020deep}:
\begin{theorem}
Let $\mathcal{L}^R$ be the function defined by eq.~(\ref{equ:ddl-main}). 
For every $\mathbf{D} \in \mathbb{R}^{m \times k}$,
let $\Qb=\Db^\top \Db+\alpha \Ib$, let  $\Fb=\Qb^{-1}\Db^\top$, and let $\cb=\Qb^{-1}\db$.
Then, for every $\mathbf{x}\in \mathbb{R}^m$,
\begin{equation}\label{equ:newMD}
   \Mb_{\Db}\mathbf{(x)}=\argmind{\ab\in\mathbb{R}^k}\mathcal{L}^R(\Db,\ab,\xb)=\operatorname{prox}^{\Qb}_{\lambda \psi}(\Fb \xb-\cb),
\end{equation}
where $\operatorname{prox}^{\Qb}_{\lambda \psi}$ denotes the proximity operator of function $\lambda \psi$ in the metric
$\|\cdot \|_{\Qb} = \sqrt{(\cdot)^\top \Qb (\cdot)}$ induced by $\Qb$ \cite{Combettes_2010,Chouzenoux14jota}.
\end{theorem}

Therefore determining the optimal sparse representation
$\mathbf{a}$ of 
$\mathbf{x}\in \mathbb{R}^m$ 
is equivalent to computing a proximity operator.
Furthermore, this result shows that the SDL can be equivalently viewed as an ADL formulation involving the dictionary matrix $\mathbf{F}$, provided that a proper metric is chosen.

\subsection{Multilayer representation}
Consequently, by substituting eq. (\ref{equ:newMD}) in eq. (\ref{equ:ddl-1}),
the DDL model can be re-expressed in a more concise and comprehensive form as
\begin{equation}\label{equ:JCDML}
    \mathbf{\hat{y}=}\varphi \circ \Ab^{(s+1)}\circ \operatorname{prox}^{\Qb^{(s)}}_{\lambda \psi_s} \circ \Ab^{(s)} \circ \dots \circ \operatorname{prox}^{\Qb^{(1)}}_{\lambda \psi_{1}} \circ \Ab^{(1)}(\xb^{(0)}),
\end{equation}
where the affine operators $\left(\Ab^{(r)}\right)_{1\leq r \leq s}$ are defined as
\begin{equation}\label{equ:transformed-ddl}
\Ab^{(r)}\colon \mathbb{R}^{k_{r-1}} \to \mathbb{R}^{k_r}\colon ~\zb^{(r-1)} \mapsto \Wb^{(r)}\zb^{(r)}-\cb^{(r)},
\end{equation}
with $k_0=m_1$, $\Wb^{(1)}=\Fb^{(1)}$, and 
\begin{equation}\label{equ:transform}
    \begin{split}
        &
        \resizebox{0.85\columnwidth}{!}{$
        \forall r\in \{2,\dots,s\},~\Wb^{(r)}=\Fb^{(r)}\Pb^{(r-1)},~\Wb^{(s+1)}=\mathbf{C} \Pb^{(s)}
        $}
        \\
        &
        \resizebox{0.78\columnwidth}{!}{$
        \forall r \in \{1,\dots,s\},\Qb^{(r)}=(\Db^{(r)})^\top \Db^{(r)}+\alpha \Ib,
        $}
        \\
        &
        \resizebox{0.78\columnwidth}{!}{$
        \Fb^{(r)}=(\Qb^{(r)})^{-1}(\Db^{(r)})^\top,~\cb^{(r)}=(\Qb^{(r)})^{-1}\db^{(r)}.
        $}
    \end{split}
\end{equation}
Eq. (\ref{equ:transformed-ddl}) shows that we obtain a structure similar to an FNN using weight operators $\left(\Wb^{(r)}\right)_{1 \leq r \leq s}$ and bias parameters $(\cb^{(r)})_{1 \leq r \leq s}$, which is referred to as the transform learning part in our DeTraMe-Net method.
In standard FNNs, the activation functions can be interpreted as proximity operators of convex functions
\cite{Combettes2018}. Eq. \eqref{equ:JCDML} evidences that our model is actually more general in the sense that different metrics are introduced for these operators. 



\subsection{Prox computation}
Reformulation \eqref{equ:JCDML} has the great advantage of exploiting algorithmic frameworks developed for FNNs provided that we are able to compute efficiently
\begin{equation}\label{equ:proximal-operator}
    \operatorname{prox}^{\Qb}_{\lambda \psi}(\Zb)=\argmind{\Ub \in \mathbb{R}^{k\times N}} \frac{1}{2}\|\Ub-\Zb\|_{F,\Qb}^2+\lambda \psi(\Ub),
 \end{equation}
where $\|\cdot\|_{F,\Qb}=\sqrt{\operatorname{tr}((\cdot)\Qb(\cdot)^\top)}$ is the $\Qb$-weighted Frobenius norm.
Hereabove, $\Zb$ is a matrix where the $N$ samples associated with the training set have been stacked columnwise. A convention is used to construct $\mathbf{X}$ and $\mathbf{Y}$ from $(\mathbf{x}_j)_{1\le j \le N}$ and $(\mathbf{y}_j)_{1\le j \le N}$. Various iterative splitting methods could be used to find the unique solution to the above convex optimization problem \cite{Combettes_2010,Komodakis2014}. Our purpose is to develop an algorithmic solution for which classical NN learning techniques can be applied in a fast and systematic manner. 
Our approach will be grounded in the following result.
\begin{theorem}\label{th:2}{\rm \cite{tang2020deep}}
Assume that an elastic-net like regularization is adopted by setting
$\psi=\|\cdot\|_1+\iota_{[0,+\infty)^{k\times N}}+\frac{\beta}{2\lambda}\|\cdot\|_{F}^2$ with $\beta \in (0,+\infty)$. For every $\Zb\in \mathbb{R}^{k\times N}$, the elements of $\operatorname{prox}^{\Qb}_{\lambda \psi}(\Zb)$ in 
eq. \eqref{equ:proximal-operator} 
satisfy 
for every $i\in \{1,\ldots,k\}$,
and $j\in \{1,\ldots,N\}$,
\begin{equation}\label{e:fixptc}
u_{i,j}=
 \begin{cases}
  \frac{q_{i,i}}{q_{i,i}+\beta}z_{i,j} - v_{i,j} & \mbox{if $q_{i,i} z_{i,j}> (q_{i,i}+\beta)v_{i,j}$}\\
  0 &\mbox{otherwise}, 
  \end{cases}
\end{equation}
where $v_{i,j}=\frac{\lambda+\sum_{\ell=1,\ell \neq i }^k q_{i,\ell}(u_{\ell,j}-z_{\ell,j})}{q_{i,i}+\beta}$.
\end{theorem}

We have observed that the choice of an elastic elastic-net like regularization
has a positive influence in increasing stability and avoiding overfitting.
Since Theorem \ref{th:2} does not provide an explicit expression of $\operatorname{prox}^{\Qb}_{\lambda \psi}(\Zb)$, we compute it by adopting a block-coordinate approach and update the $i$-th
row of $\mathbf{U}$ by fixing all the other ones. As $\mathbf{Q}$ is a positive definite matrix, $q_{i,i} > 0$.
Let 
\begin{equation*}\label{e:reparam}
\begin{split}
    &
    \resizebox{0.99\columnwidth}{!}{$
    \widetilde{\mathbf{W}}= - 
    \left(\frac{q_{i,\ell}}{q_{i,i}+\beta}\delta_{i-\ell}\right)_{1\leq i,\ell \leq k},~\mathbf{h}=\left(\frac{q_{i,i}}{q_{i,i}+\beta}\right)_{1\leq i \leq k}\in [0,1]^k,
    $}
    \\
    &\resizebox{0.99\columnwidth}{!}{$\mathbf{b}=\left(\frac{\lambda}{q_{i,i}+\beta}\right)_{1\leq i \leq k}\in [0,+\infty)^k,~\mathbf{1}=[1,\ldots,1]^\top\in \mathbb{R}^N,$}\\
\end{split}
\end{equation*}
where $(\delta_\ell)_{\ell \in \mathbb{Z}}$ is the Kronecker sequence
(equal to 1 when $\ell=0$ and 0 otherwise).
Then, \eqref{e:fixptc} suggests that the elements of
$\Ub$ can be globally updated, at iteration $t$,
as follows
\begin{equation}\label{equ:u-update}
\Ub_{t+1}=
  \operatorname{ReLU}\big((\mathbf{h} \mathbf{1}^\top)\odot \Zb+\widetilde{\Wb}(\Ub_t-\Zb)-\mathbf{b} \mathbf{1}^\top\big),  
\end{equation}
where $\odot$ denotes the Hadamard (element-wise) product.
Note that a similar expression can be derived by applying a preconditioned 
forward-backward algorithm \cite{Chouzenoux14jota} to Problem \eqref{equ:proximal-operator}, where the 
preconditioning matrix is $\operatorname{Diag}(q_{1,1},\ldots,q_{k,k})$.
Given $\widetilde{\mathbf{W}}$, $\mathbf{h}$, and $\mathbf{b}$,
this updating rule \eqref{equ:u-update} can be viewed as an RNN structure
for which $\Ub_t$ is the hidden variable and $\Zb$ is a constant input over time. 

\renewcommand{\algorithmicrequire}{\textbf{Initialization:}}
\renewcommand{\algorithmicensure}{\textbf{Output:}}
{\small
\begin{algorithm}[h!] 
	\caption{Deep Metric and Transform Learning }  
	\label{alg:alg2}  
	\begin{algorithmic}[1]
	    \REQUIRE Set $t=0$.
		\WHILE {not converged \text{and} $t< t_{\rm max}$}
		\STATE \textbf{Forward pass: $\mathbf{U}_t^{(0)}=\mathbf{X}$}
		\FOR{$r=1,\dots,s+1$}
		\STATE $\mathbf{Z}_t^{(r)}=\mathbf{W}_t^{(r)}\mathbf{U}_t^{(r-1)}-\mathbf{c}_t^{(r)}$
		\IF{$r\leq s$} 
        \STATE Initialize $\mathbf{U}^{(r)}_0$ as the null matrix and set $tt=0$
		\WHILE {not converged \text{and} $tt< tt_{\rm max}$}
		\STATE 
		Update $\mathbf{U}_{tt+1}$ according to eq. (\ref{equ:u-update}); $tt\leftarrow tt+1$
		\ENDWHILE
		\ENDIF
		\ENDFOR
		\STATE $\mathbf{\hat{Y}}_t = \varphi(\mathbf{Z}_t^{(s+1)})$
		\STATE \textbf{Loss:} \mbox{$\mathcal{L}'(\boldsymbol{\theta}_t) = \mathcal{L}(\mathbf{Y},\mathbf{\hat{Y}}_t)$, $\boldsymbol{\theta}_t$: vector of all parameters}
		\STATE \textbf{Backward pass:}
		\FOR{$r=1,\dots,s+1$}
		\STATE $\mathbf{W}_{t+1}^{(r)}=\mathbf{W}_t^{(r)}-\rho_t \frac{\partial \mathcal{L}'}{\partial \mathbf{W}^{(r)}}(\boldsymbol{\theta}_t)$ 
		\STATE$\mathbf{c}_{t+1}^{(r)}=\mathbf{c}_t^{(r)}-\rho_t \frac{\partial \mathcal{L}'}{\partial \mathbf{c}^{(r)}}(\boldsymbol{\theta}_t)$
		\ENDFOR
		\FOR{$r=1,\dots,s$}
		\STATE $\widetilde{\mathbf{W}}_{t+1}^{(r)}=\mathsf{P}_{\mathcal{D}_0}\left(\widetilde{\mathbf{W}}^{(r)}_t-\rho_t \frac{\partial \mathcal{L}'}{\partial \widetilde{\mathbf{W}}^{(r)}}
		(\boldsymbol{\theta}_t)\right)
		$
		\STATE $\mathbf{h}_{t+1}^{(r)}=\mathsf{P}_{[0,1]^k}\left(\mathbf{h}_t^{(r)}-\rho_t \frac{\partial \mathcal{L}'}{\partial \mathbf{h}^{(r)}}(\boldsymbol{\theta}_t)\right)$
		\STATE$\mathbf{b}_{t+1}^{(r)}=\mathsf{P}_{[0,+\infty)^k}\left(\mathbf{b}_t^{(r)}-\rho_t \frac{\partial \mathcal{L}'}{\partial \mathbf{b}^{(r)}}(\boldsymbol{\theta}_t)\right)$
		\ENDFOR
		\STATE $t\leftarrow t+1$
		\ENDWHILE
	\end{algorithmic}  
\end{algorithm} 
}

\begin{table*}[t]
    \centering
    \scriptsize{
    \begin{tabular}{c|c|c|c|c}
    \hline
    \hline
    Accuracy (\%)  &\multicolumn{2}{|c|}{CIFAR10 +} &\multicolumn{2}{|c}{CIFAR100 +}\\
    \hline
         Network Architectures   & Original  & DeTraMe-Net & Original  & DeTraMe-Net \\
         &  & (\#iteration) &    & (\#iteration)  \\
         \hline
         PlainNet 3-layer & 35.14 $\pm$ 4.94 & \textbf{88.51} $\pm$ 0.17 (5)&22.01 $\pm$ 1.24 & \textbf{64.99} $\pm$ 0.34 (3)\\
         PlainNet 6-layer& 86.71 $\pm$ 0.36 & \textbf{92.24} $\pm$ 0.32 (2)&62.81 $\pm$ 0.75&\textbf{69.49} $\pm$ 0.61 (2)\\
         PlainNet 9-layer  &90.31 $\pm$ 0.31 & \textbf{93.05} $\pm$ 0.46 (2)&66.15 $\pm$ 0.61 &\textbf{69.68} $\pm$ 0.50 (2)\\
         PlainNet 12-layer &91.28 $\pm$ 0.27 & \textbf{92.03} $\pm$ 0.54 (2)&68.70 $\pm$ 0.65 &\textbf{70.92} $\pm$ 0.78 (2)\\
         \hline
         ResNet 8 &87.36 $\pm$ 0.34 & \textbf{89.13} $\pm$ 0.23 (3)&60.38 $\pm$ 0.49 &\textbf{64.50} $\pm$ 0.54 (2)\\
         ResNet 20 &92.17 $\pm$ 0.15 & \textbf{92.19} $\pm$ 0.30 (3)&68.42 $\pm$ 0.29 &\textbf{68.62} $\pm$ 0.27 (2)\\
         ResNet 56 &93.48 $\pm$ 0.16 & \textbf{93.54} $\pm$ 0.30 (3)&\textbf{71.52} $\pm$ 0.34 &\textbf{71.52} $\pm$ 0.44 (2)\\
         ResNet 110 &93.57 $\pm$ 0.14& \textbf{93.68} $\pm$ 0.32 (2)&72.99 $\pm$ 0.43 &\textbf{73.05} $\pm$ 0.40 (2)\\
         \hline
        \hline
    \end{tabular}
    \caption{
    {
    \small{CIFAR10 and CIFAR100 with + is trained with simple translation and flipping data augmentation. All the presented results are re-implemented over 5 runs by using the same settings, and are calculated by their means and standard deviations.}
    }\label{tab:results}
    }
    }
\end{table*}

\subsection{Training procedure}
We have finally transformed our DDL approach in an alternation of
linear layers and specific RNNs. 
Let $\rho_t>0$ be the learning rate at iteration $t$, the simplified 
form of a training method for DeTraMe-Net is provided in Alg.~\ref{alg:alg2}.
The constraints on the parameters of the RNNs have been imposed by projections.
In Alg.~\ref{alg:alg2}, $\mathsf{P}_{\mathcal{S}}$ denotes the projection onto a nonempty closed convex set
$\mathcal{S}$ and
$\mathcal{D}_0$ is the vector space of $k\times k$ matrices with diagonal terms equal to 0.

\section{Experiments and Results}
\label{sec:experiments}
In this section, our DeTraMe-Net method is evaluated on two popular datasets, namely CIFAR10 \cite{CIFAR} and CIFAR100 \cite{CIFAR}. Since the common NN architectures are plain networks such as ALL-CNN \cite{ALLCNN} and residual ones, such as ResNet \cite{resnet}
, we compare DeTraMe-Net with these two respective state-of-the-art architectures. 
We replace all the ReLU activation layers in PlainNet with Q-Metric ReLU, leading to DeTraMe-PlainNet, and replace the ReLU layer inside the block in ResNet by Q-Metric ReLU, giving rise to DeTraMe-ResNet. More detailed information can be found in \cite{tang2020deep}.
\subsection{DeTraMe-Net vs. DDL}
\begin{table}[htb!]
    \centering
    \small{
    \resizebox{\columnwidth}{!}{
    \begin{tabular}{|c|c|c|c|}
    \hline
         Model & \# Parameters &CIFAR10 &CIFAR100  \\
    \hline
         DDL 9 \cite{mahdizadehaghdam2019deep} &1.4M& 93.04\%$^*$ & 68.76\%$^*$ \\
         DeTraMe-Net 9&3.0M&\textbf{93.05\%} $\pm$ 0.46\%  &\textbf{69.68\%} $\pm$ 0.50\%  \\
    \hline
    \hline
         Model& \#Parameters  & Training (s/per image)   &Testing (s/per image)  \\
    \hline
         DDL \cite{mahdizadehaghdam2019deep}& {0.35 M}&  0.2784$^*$ & 9.40$\times 10^{{-2} ^*}$ \\
         DeTraMe-Net 12&\textbf{2.4 M}&\textbf{0.1605} &\textbf{3.52}$\times 10^{-4} $  \\
         
    \hline
    \end{tabular}}
    \caption{\small{The results with $^*$ are reported in the original paper. The top table provides the accuracy of DeTraMe-Net vs. DDL, where DeTraMe-Net 9 and DDL 9 follow the same ALL-CNN \cite{ALLCNN} architecture. The bottom one is the efficiency comparison, where the architectures are different from the top ones
    }}\label{tab:ddlcompare}
    }
\end{table}

         

First we compare our DeTraMe-Net with DDL \cite{mahdizadehaghdam2019deep}. Although DeTraMe-Net needs more parameters, DeTraMe-Net presents two main advantages: (1) A better ability to discriminate: in comparison to DDL in the top table of Table \ref{tab:ddlcompare}, in terms of averaged performance, $0.01\%$ and $0.92\%$ accuracy improvements are respectively obtained on these two datasets. (2) DeTraMe-Net is implemented in a network framework, with no need for extra functions to compute gradients at each layer, which greatly reduces the time costs. 
The bottom part of Table \ref{tab:ddlcompare} shows that our method with 6 times more parameters than DDL only requires half training time and a faster testing time by a factor 100. Moreover, by taking advantage of the developed implementation frameworks for neural networks, DeTraMe-Net can use up to 110 layers, while the maximum number of layers in \cite{mahdizadehaghdam2019deep} is 23.
\subsection{DeTraMe-Net vs. Generic CNNs}
We next compare DeTraMe-Net with generic CNNs with respect to two different aspects: \underline{Accuracy} and \underline{Capacity}.

\textbf{Accuracy.} As shown in Table \ref{tab:results}, 
with the same architecture, using DeTraMe-Net structures achieves an overall better performance than all various generic CNN models do. For PlainNet architecture, DeTraMe-Net increases the accuracy with a median of $3.99\%$ on CIFAR10, $5.11\%$ on CIFAR100 and $0.45\%$ on SVHN.
For ResNet architecture, DeTraMe-Net also consistently increases the accuracy with a median of $0.05\%$ on CIFAR10, $0.13\%$ on CIFAR100.

\textbf{Capacity.}  
In terms of \textit{depth}, comparing improvements with PlainNet and ResNet, 
shows that the shallower the network, the more accurate. It is remarkable that DeTraMe-Net leads to more than $42\%$ accuracy increase for PlainNet 3-layers 
on CIFAR10 and CIFAR100. 
When the networks become deeper, they better capture discriminative features of the classes, 
and albeit with smaller gains, 
DeTraMe-Net still achieves a better accuracy than a generic deep CNN, e.g. around $0.11\%$ and $0.05\%$  higher than ResNet 110 on CIFAR10 and CIFAR100. 

\section{Conclusion}
\label{sec:conclusion}
Starting from a DDL formulation, we have shown that the introduction of metrics within standard activation operators, gives rise to a novel Joint Deep and Transform Learning problem.
This has allowed us to show that the original DDL can be performed thanks to a network mixing FNN and RNN parts,
so leading to a fast and flexible network framework for building efficient DDL-based classifiers.
Our experiments illustrate that the resulting DeTraMe-Net performs better than the original DDL approach
and state-of-the-art generic CNNs. 




\fontsize{9}{9}\selectfont
{
\bibliographystyle{IEEEbib}
\bibliography{refs}
}
\end{document}